\documentclass[conference]{IEEEtran}
\IEEEoverridecommandlockouts
\usepackage{subcaption}

\usepackage[ruled,linesnumbered]{algorithm2e}
\usepackage{amsmath,amssymb,amsfonts}
\usepackage{graphicx}
\usepackage{soul,color}
\usepackage{authblk}
\usepackage{booktabs}

\setstcolor{red}

\usepackage[table]{xcolor} % Add xcolor package for coloring tables

\begin{document}

\title{A Privacy-Preserving Indoor Localization System based on Hierarchical Federated Learning
 }

\author{Masood Jan, Wafa Njima, Xun Zhang
\thanks{This work was supported by the European Union’s Horizon 2020 research and innovation program under grant agreement No 101017226, project 6G BRAINS.}}

\affil{ISEP, Institut Sup\'{e}rieur d'Electronique de Paris, 75006 Paris, France  \authorcr Emails: \{masood.jan, wafa.njima, xun.zhang\}@isep.fr}


\maketitle

% \pagestyle{plain}
% \fancyfoot{}

\begin{abstract}

Location information serves as the fundamental element for numerous Internet of Things (IoT) applications. Traditional indoor localization techniques often produce significant errors and raise privacy concerns due to centralized data collection. In response, Machine Learning (ML) techniques offer promising solutions by capturing indoor environment variations. However, they typically require central data aggregation, leading to privacy, bandwidth, and server reliability issues. To overcome these challenges, in this paper, we propose a Federated Learning (FL)-based approach for dynamic indoor localization using a Deep Neural Network (DNN) model. Experimental results show that FL has the nearby performance to Centralized Model (CL) while keeping the data privacy, bandwidth efficiency and server reliability. %We also compared our FL model results with the CL model on KNN approach to see the differences in the evaluation matrix. 
This research demonstrates that our proposed FL approach provides a viable solution for privacy-enhanced indoor localization, paving the way for advancements in secure and efficient indoor localization systems.
\end{abstract}


\section{Introduction}
Most of the Internet of Things (IoT) applications such as emergency services, e-marketing and social networking critically exploit location information to operate properly \cite{tsai2017precise}.
Traditional global positioning system (GPS) has revolutionized the outdoor navigation \cite{hinch2010outdoor}, however, it often fell short in indoor environments because of the weak signal strength, poor penetration and lack of coverage. %\cite{wu2020using}. 
Therefore, we need an efficient and robust indoor advanced localization system to insure consistent performance in challenging indoor environments. In fact, multi-path effects caused by the high dynamics of indoor environments can affect the accuracy of traditional localization methods resulting in significant localization errors \cite{njima2020deep}.

In this regard, numerous cutting-edge methods for target localization, target tracking, and navigation
were introduced \cite{gustafsson2005mobile}, \cite{sayed2005network}. However, in complicated indoor environments (i.e., offices, malls, museums, etc.), the majority of existing works may seriously mismatch the underlying mechanism since they rely on empirical, parametric transitions and measurement models, which can be thought of as an individual abstract of human experience. Such a model mismatch may be mitigated and the location accuracy can be improved by learning from a large amount of historical data \cite{yin2020fedloc}. For this, the integration of machine learning (ML) techniques has been recognized as essential in advancing indoor localization models \cite{nessa2020survey}. % mainly deep neural networks (DNNs) \textcolor{red}{You have to justify why DNN or keep only ML without talking about DNN.}. 
%Through continuous training, we can develop and optimize a ML model such as a deep neural network (DNN). 
Once trained, the ML model is capable of estimating accurate predictions or decisions. 
In traditional machine learning frameworks, models are trained within a centralized learning network (CLN), where a server gathers and retains all training data in a centralized dataset \cite{tan2020toward}. However, with the rapid advancements in communication technologies and integration of digital services and industries, the volume of data generated within indoor environments has surged exponentially. Sending such a vast amount of data to a central server poses several challenges and concerns. 

In fact, the centralized approach raises data security and privacy risks, as sensitive information may be vulnerable to breaches or unauthorized access during transmission and storage. For instance, during the COVID-19 pandemic, the need to share location data to monitor the spread of infections and identify high-risk areas became crucial. However, this also highlighted significant privacy concerns, as the transmission of location data to central servers increased the risk of unauthorized access to individuals' movements and personal information, underscoring the urgent need to address mobile users' location privacy \cite{greenberg2020apple}. 
Also, the sheer volume of data being transmitted over the network imposes significant strain on available bandwidth, potentially leading to network congestion and degraded performance. 

To address these critical issues, the adoption of a Federated Learning (FL) based approach \cite{nguyen2021federated} becomes increasingly appealing. With this model, the data remains localized and confined within each device, thereby significantly reducing the exposure of sensitive information to potential threats. In addition, the local processing and model training conducted by FL enables reduced data transmission, as only the model updates (such as gradients or updated model parameters) are transmitted between devices and the central server, rather than the raw data itself. This enhances privacy by minimizing the need to share sensitive data and alleviates the bandwidth burden on the communication infrastructure. Moreover, the implementation of FL aligns well with the emerging emphasis on energy efficiency and resource optimization. %These advantages make the FL a compelling and promising approach to address the challenges posed by dynamic indoor environments and pave the way for more secure indoor positioning solutions in our increasingly connected digital world \cite{de2023hed}.

In recent years, different localization works based on FL have been published.% Authors in \cite{mcmahan2017communication} introduced FL for the first time in 2017. 
  Several recent works include the simultaneous use of FL and Deep Neural Network (DNN). In \cite{kumar2023confidentiality}, a federated architecture is proposed for Wi-Fi fingerprinting that maintains data privacy across edge devices while enhancing localization accuracy, however this approach trains a model for each building, posing privacy risks in larger scenarios like smart cities where each building contains substantial data. Another approach in \cite{etiabi2023federated} collects data from all buildings, shuffles it, and then creates clients from this shuffled data to better tune the model for each client. This method compromises the fully privacy-preserving nature of FL .

In this paper, we propose a decentralized model training approach by adapting a three-tier FL model: local server training at the floor level, regional server training at the building level (with only the floor model weights), and global server training at the regional level (with only the regional model weights). Thus, raw data remains local to each floor, ensuring it is not transmitted to central or regional servers and only the trained model weights are shared, enhancing privacy compared to previous works. 
To validate the effectiveness of our approach, we conducted extensive performance evaluations using the widely recognized UJIIndoorLoc database \cite{torres2014ujiindoorloc} which contains real recorded data of received signal strength indicator (RSSI) measurements, from different Access Points (APs) deployed in a campus indoor environment, associated to corresponding positions' coordinates. We extended the code to develop a centralized machine learning model for direct computation on the same database in order to compare and analyze the results obtained from both models and to gain insights into their respective performances. Furthermore, we benchmarked our FL model against a Centralized Learning (CL) model using the K-Nearest Neighbors (KNN) algorithm, highlighting the superior performance of our FL framework in real-world indoor localization scenarios.

The rest of the paper is organized as follows:
% Section~\ref{sec:related-work} deals with the related works while
Section~\ref{sec:hdnn} presents the system model followed by Section~\ref{algorithm} which details the hierarchical learning scheme, the data preprocessing and the algorithm development of our localization problem. Section~\ref{sec:Comparison} depicts the comparison with the state of the art learning methods. In Section~\ref{sec:results}, we go through the performance evaluation and analysis of our model before concluding in Section~\ref{sec:Conclusion}.

\section{System model}
\label{sec:hdnn}

%First, we will train a local floor model for each floor, then we use the weights of these floor models to train a regional building model. Then we aggregate the weights of these building models to train a FL global model. Finally, we will choose the same model architecture for the Centralized Central Model as used for the FL Central Model to evaluate the performance on a separate validation dataset and obtain unbiased performance estimates.

%\subsection{Hierarchy learning scheme of FL infrastructure}

\begin{figure*}[!t] % Use the asterisk (*) to span both columns
  \centering
  \includegraphics[width=\linewidth, height = 9.5cm]{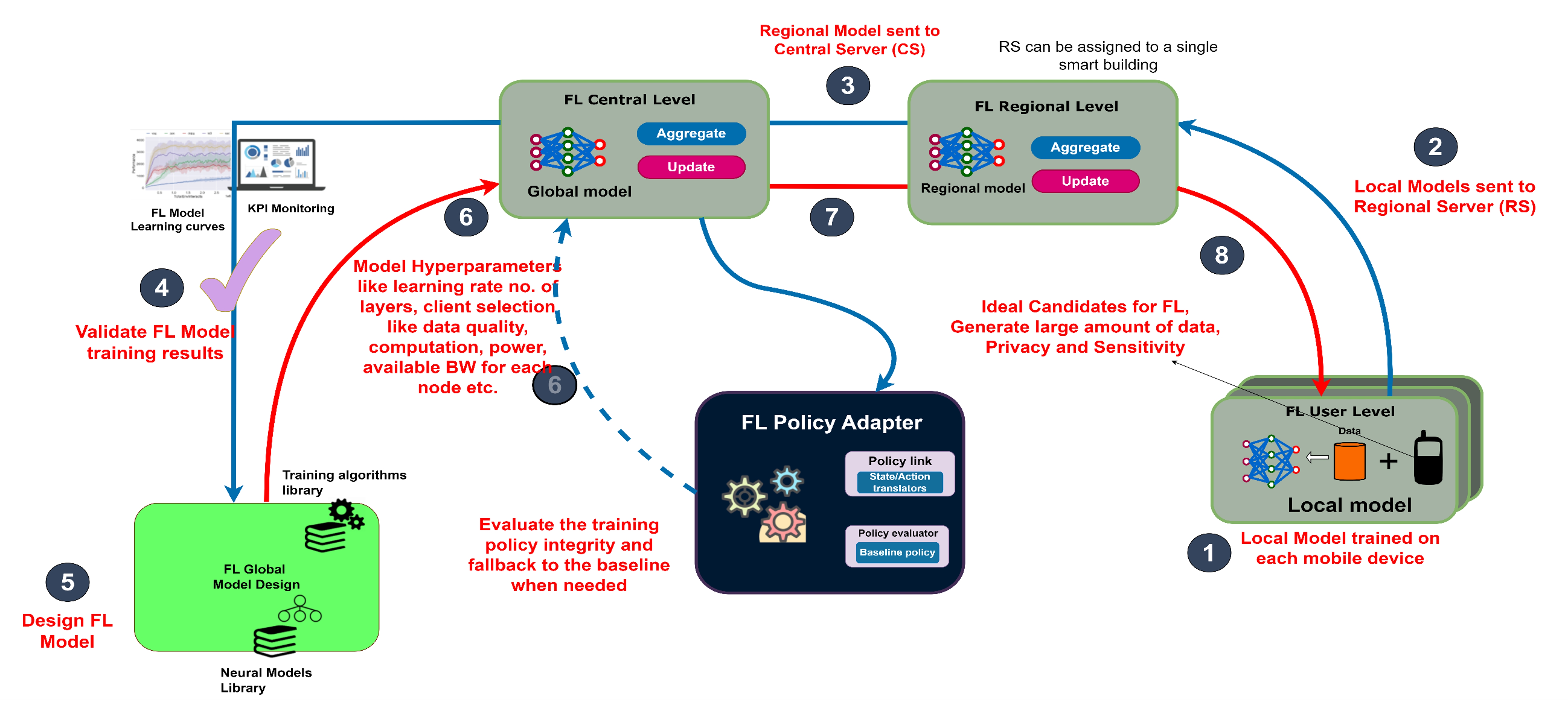}
  \caption{Hierarchical representation of FL based localization process.} % Add a caption for your image
  \label{fig:1} % Add a label for referencing the image in the text
\end{figure*}

Our goal is to implement FL-based approach to predict the users' coordinates (longitude and latitude) indoor based on collected Wi-Fi RSSI data. A common representation of FL is expressed as follows: 
consider a global loss function, denoted as $\mathcal{L}$ which is derived from a weighted combination of $\mathcal{K}$ local losses $\{ {\mathcal{L}}_k \}_{k=1}^{K}$. These local losses are computed from private data $X_k$ which remains exclusively with the respective parties involved and is never shared between them \cite{rieke2020future}.
\begin{equation}
\min_{\phi} \mathcal{L}(X;\phi) \quad \text{with} \quad \mathcal{L}(X;\phi) = \sum_{k=1}^{K} w_k \mathcal{L}_k(X_k;\phi),
\end{equation}
where $w_k>0$ denote the respective weight coefficients.
Fig. 1 illustrates the hierarchical representation of our FL-based localization process.
We have considered a multi-building and multi-floor indoor environment containing $N$ buildings $(B_0, B_1,..., B_N)$ and each building contains $M$ floors $(F_0, F_1,..., F_M)$. 
Each floor has its own set of Wi-Fi access points (APs). We considered each floor as a single user due to the limitation of data. Each floor from each building trains its own floor model using its local Wi-Fi APs and the associated RSSI data. This is done independently on each floor without sharing the raw data with a central server. Each floor model aims to predict the longitude and latitude coordinates of locations inside its corresponding floor. 

Once the floor models are trained, the weights of these models are aggregated to create building-level models, referred to as regional models. For each building, the floor models' weights are combined to create a regional model that represents the collective knowledge of that building. The regional models from different buildings $(B_0, B_1,..., B_N)$ are further aggregated to create the FL central model. This central model is the global model that predicts the longitude and latitude coordinates of locations inside all the buildings.
Our Key Performance Indicator (KPI) is the Mean Absolute Error (MAE) which is monitored and validated using the training results of the FL central model. This step is essential to ensure the model's satisfying performance. Furthermore, the model hyper parameters mainly the learning rate, the number of layers, and the floor selection (data quality, computation, etc.) can be adjusted to optimize the FL central model's performance.

Once the FL central model is trained and evaluated, its updated weights are shared back with the regional models. This allows each building to benefit from the knowledge learned across the buildings without sharing sensitive data centrally. Additionally, the local floor models can also receive updates from the FL central model, enhancing their prediction capabilities. This process of aggregating and sharing weights between floor models, regional models, and the FL central model will go through multiple rounds of training and updating to continually improve their performance. The update frequency of the model is determined based on the rate of data variation and the need of the considered localization service.

\section{Hierarchy learning scheme of FL infrastructure for indoor localization}\label{algorithm}

\subsection{Data preprocessing}

 As said before, we apply our model to the UJIIndoorLoc database presented in \cite{torres2014ujiindoorloc} which is the first and largest publicly available real use case database. We have three buildings ($N = 3$) and each building is composed of four floors ($M = 4$). The distribution of collected data is presented in TABLE~\ref{Table:z}. In this table, we mention the number of collected fingerprints corresponding to each floor and its percentage compared to the whole collected data. 
\begin{table*}[ht]
\caption{Data distribution per floor per building.}
\centering
\scalebox{1.2}{
\begin{tabular}{|c|c|c|c|c|c|c|c|c|c|c|}
\hline
 & \multicolumn{2}{c|}{Floor 0} & \multicolumn{2}{c|}{Floor 1} & \multicolumn{2}{c|}{Floor 2} & \multicolumn{2}{c|}{Floor 3}\\
\hline
Building 0 & 14464 & 4\% & 22700 & 7\% & 26146 & 8\% & 21670 & 6\% \\
\hline
Building 1 & 23197 & 7\% & 27609 & 8\% & 26582 & 8\% & 10172 & 3\% \\
\hline
Building 2 & 32452 & 10\% & 47169 & 14\% & 35381 & 10\% & 53819 & 16\% \\
\hline
\end{tabular}
}
\label{Table:z}
\end{table*}
This database contains the RSSI measurement from multiple APs deployed in the considered campus indoor setting. Each AP represent one column in the dataset, so there will be 520 columns showing the measurements from 520 APs. The database has a training dataset which contains 19937 RSSI measurements and a validation dataset with 1111 RSSI measurements. Note that the validation dataset is measured 4 months after the training dataset to adapt a real-world use scenario. During the measurement process some APs readings fell beyond the range of the user device and these values are represented by 100 dBm. 
 
We start our preprocessing by deleting all those columns where there is no data recorded at all i-e; cell values with 100 dBm. This step reduces the total number of columns to 465 APs. Then we take out those APs (columns) whose contribution to the dataset’s richness is insignificant. We establish a threshold for APs at $\tau =0.98$. This means that if an AP is missing in over 98\% of the measurements, it is considered insignificant to the dataset's richness and is therefore excluded. After applying this criterion, we retain only 248 APs.

In the next step, we address the 100 dBm values in the dataset as missing RSSI values, indicating no signal from the respective access point at those locations. Thus, we can treat these 100 dBm values as the overall minimum RSSI values denoted by $\text{\textit{min}}_{\scriptstyle \text{\textit{RSSI}}}$. Since the lowest recorded value in the dataset is -104 dBm, we assign the $\text{\textit{min}}_{\scriptstyle \text{\textit{RSSI}}} = -105$ dBm. The choice of -105 dBm as the $\text{\textit{min}}_{\scriptstyle \text{\textit{RSSI}}}$ threshold was made to ensure uniform treatment of missing data, with the value being slightly lower than the lowest observed RSSI value of -104 dBm.

Next we implement the method explained in \cite{torres2015comprehensive}, which involves converting the RSSI measurements into a powered representation by setting the exponent $\beta$ to $e$. This transformation is required to represent the data in normalized and positive values, which enhances the performance of the DNN. This representation is defined as;

\resizebox{0.9\linewidth}{!}{
\begin{minipage}{\linewidth}
\begin{equation}
    powed\left(RSSI_i\right) = \left(\frac{RSSI_i - min_{RSSI}}{-min_{RSSI}}\right)^\beta,
\end{equation}
\end{minipage}
}

\textit{powed}($RSSI_i$) is the output value of the equation which represent the transformation applied to the input RSSI value. \textit{$RSSI_i$} is the input value, representing RSSI at the specific instance. $\text{\textit{min}}_{\scriptstyle \text{\textit{RSSI}}}$ is the minimum possible value of RSSI in the dataset. It's the value that \textit{$RSSI_i$} is being compared against. $\beta$ is a parameter controlling the exponentiation. 

The performance of our model is assessed through the MAE, the training losses and validation losses. The MAE refers to the Mean Absolute Error and can be defined as follows:
\begin{equation}
\resizebox{.9\hsize}{!}{${2D-MAE}=\frac{1}{N} \sum_{i=1}^{N}\left [   \delta \left(\mathcal{B}_i, \hat{ \mathcal{B}_i}\right)\cdot \delta \left( \mathcal{F}_i, \hat{ \mathcal{F}_i}\right)\cdot\sqrt{\left ( x_i-\hat{x}_i \right )^2+\left ( y_i-\hat{y}_i \right )^2}\ \right ]$},
\end{equation}

Where $\sigma$(a,b) = 1, if a = b, 0 otherwise. N is the total number of estimated positions. $\beta_i$ and $\hat\beta_i$ are actual and predicted attributes, such as signal strengths or environmental factors, while $F_i$ and $\hat{F_i}$  are actual and predicted features, representing additional conditions or characteristics relevant to the $i^{th}$ term in the summation. $x_i$, $\hat{x}_i$, $y_i$, and $\hat{y}_i$ are the real and estimated coordinates values, respectively. In the context of our study, the MAE is measured in meters. This unit reflects the average distance between the predicted and actual geographic coordinates, thus providing a direct indication of the accuracy of our localization model.

\subsection{Algorithm Development}

Our DNN model is constructed using the TensorFlow Keras API, providing a flexible and scalable framework for building neural network architectures \cite{chicho2021comprehensive}. The model is designed with the objective of capturing complex relationships within the data and enabling accurate predictions in a decentralized and collaborative environment. %\cite{aldoseri2023re}.

The optimized model architecture used for this project has three hidden layers and a linear output layer. The activation function used in the hidden layers is Rectified Linear Unit (ReLU), a non-linear activation function that helps neural networks capture complex relationships and overcome gradient-related issues. 
The final output layer used a linear activation function as well. 

To optimize the DNN model during training, we employ a customized error function that aligns with the learning objectives and dataset characteristics. The error function used in our federated learning setup is the Mean Absolute Error (MAE). The configuration of our model which will be applied is depicted in TABLE~\ref{Table:a}. Note that for fair comparison, we will use the same configuration for our FL model as well as for the CL model. The simulation settings for our FL approach are provided in TABLE~\ref{Table:b}.

\begin{table*}[ht]
\caption{DNN Models Configuration.}
\centering
\scalebox{1.2}{
\begin{tabular}{|c|c|c|c|c|c|}
\hline
\multicolumn{2}{|c|}{\textbf{Floor Model (Local Model) }} & \multicolumn{2}{c|}{\textbf{Building Model (Regional Model)}} & \multicolumn{2}{c|}{\textbf{Global Model}} \\ \hline
    hidden layers & 256 - 64 & hidden layers & 256 - 64 & hidden layers & 256 - 64 \\ \hline
    dropout layer & 0.25 - 0.1 & dropout layer & 0.25 - 0.1 & dropout layer & 0.25 - 0.1 \\ \hline
    input activation & ReLU & input activation & ReLU & input activation & ReLU \\ \hline
    output activation & linear & output activation & linear & output activation & linear \\ \hline
  \end{tabular}
  }
  \label{Table:a}
\end{table*}

\begin{table}[ht]
\caption{FL Simulation Settings.}
    \centering
    \scalebox{1}{
    \begin{tabular}{ccc}
        \toprule
        \textbf{Parameter} & \textbf{Description} & \textbf{Value} \\
        \midrule
        Optimizer & Model optimizer & Adam \\
        $\eta$ & Learning rate & 0.0005 \\
        $\beta 1$, $\beta 2$ & Exponential decay rates & 0.1, 0.99 \\
        \textit{C} & Number of Users (Floors) & 12 \\
        \textit{B} & Batch size & 32 \\
        \textit{E\textsubscript{1}} & Number of epochs per local iteration & 10 \\
        \textit{E\textsubscript{2}} & Number of epochs per central iteration & 1000 \\
        \textit{R} & Communication rounds & 100 \\
        \bottomrule
    \end{tabular}
    }
    \label{Table:b}
\end{table}

We design our algorithm to take as input a set of user RSSI datasets denoted as $D_c$ and a set of output weights $\alpha u$. The trained model parameters, represented as $W$ will be the output. The algorithm starts by initializing the server. It sets the initial model parameters $W_0$ and a variable $r$ to track the communication rounds. Then the algorithm enters a loop that continues until either convergence is reached or a maximum number of communication rounds (max\_com\_rounds) is exceeded. In each communication round $r$, the server broadcasts the current global model $W_r$ to all participating users. For each floor $F$, the algorithm performs parallel local training using the following steps:

- The local training process on a specific floor

- The local model on that floor gets updated with the current global model $W_r$
     
- The local model is trained using gradient descent with learning rate $\mu$ and gradient $\nabla F_{W_{r,k}^c} (\beta_c)$, where $W_{r,k}^c$ is the local model on the $k$-th iteration for the $c$-th user on floor $F$

- The updated local model parameters are represented as $W_{r}^c$ and are uploaded by the floor.

After all floors have performed their local training, each building aggregates the local models of its floors to train a regional model. This regional model represents the collective knowledge of all floors within the building.

The server then performs a global update step using the uploaded regional models and the corresponding dataset sizes $D_c$. The global model parameters $W_{r+1}$ are updated as a weighted average of the regional models:
\begin{equation}
W_{r+1} = \frac{1}{\sum_{c}^{|D_c|}} \sum_{c} |D_c| W_{r}^c,
\end{equation}

The variable $r$ is incremented to move to the next communication round.

Finally, the algorithm checks whether the convergence criterion has been met or if the maximum communication rounds have been reached. If neither condition is met, the loop iterates again. The entire process of Federated Learning is described by Algorithm 1.

\begin{algorithm}
  \caption{Federated Learning for Localization.}
  \SetAlgoLined
  \SetKwInOut{Input}{Input}
  \SetKwInOut{Output}{Output}
  
  \Input{\{Dc\}; /* User RSSI Datasets */}
  \Input{$\{\alpha u\}$; /* Outputs weights */}
  \Output{\{W\}; /* Trained model parameters */}
  
  \BlankLine
  \SetKwProg{ServerInit}{ServerInit}{}{}
  \ServerInit{()}{
    Set $W_0$, and $ r \gets 0$ \;
  } 

  \While{not converged and $r < \text{max\_com\_rounds}$}{
    Server broadcasts $W_r$\;
  
    \ForEach{Floor $F \in \{1, 2, \ldots, F\}$}{
      /* In parallel */
  
      \SetKwProg{FloorLocalTraining}{FloorLocalTraining}{}{}
      \FloorLocalTraining{$(\mathcal{D}_c, \alpha_u, W_r$)}{
        Update local model with $W_r$ \;
        Train local model using
        $
        \mathbf{W}_{r,k+1}^c = \mathbf{W}_{r,k}^c - \mu\nabla F_{\mathbf{W}_{r,k}^c}(\mathcal{B}_c)
        $ \;
        Floor uploads $W_r^c$ \;
      }
    }
  
    /* Aggregate floor models to create building models (regional models) */
    \ForEach{Building $B \in \{1, 2, \ldots, N\}$}{
      \SetKwProg{RegionalModelTraining}{RegionalModelTraining}{}{}
      \RegionalModelTraining{$(\{W_r^c, \left|\mathcal{D}_c\right|\})$}{
        Aggregate floor models' weights belonging to building $B$ to create regional model weights for building $B$ \;
        Train regional model for building $B$ using the aggregated weights \;
      }
    }
  
    /* Combine regional models to update global model */
    
    \SetKwProg{GlobalModelTraining}{GlobalModelTraining}{}{}
    \GlobalModelTraining{$(\{W_r^c, \left|\mathcal{D}_c\right|\})$}{
      Aggregate regional models' weights to create global model weights \;
      Train global model using the aggregated weights \;
    }
  
    $ r \gets r +1$ \;
  }
\end{algorithm}

\section{Comparison with state of the art methods}
\label{sec:Comparison}
After training the FL central model, we train a CL central model using all the data from the buildings without the privacy constraints of FL and consider them as a single dataset. We choose the same model architecture for the CL central model as used for the FL central model. The goal is to compare the performance of the CL model with the FL model as a benchmark to assess how well the FL approach generalizes to a centralized scenario. The model architecture should be consistent to ensure a fair comparison between the two approaches. During training, the model is updated using back propagation and gradient descent to minimize the prediction errors. %\cite{zhao2023cascaded}. 
After training both models, we evaluate the performance of both models on a separate validation dataset to obtain unbiased performance estimates. 

%Lastly, we assess the evaluation metrics for both models and see whether the FL approach is able to achieve comparable or better results compared to the traditional centralized approach while maintaining data privacy and security.

%It's worth noting that the performance of the FL central model might not be exactly the same as the CL central model due to factors like communication overhead, varying data distributions across buildings, and potential limitations in communication bandwidth \cite{asad2023limitations}. However, the goal of FL is not necessarily to achieve identical results but to come close to the centralized performance while ensuring data privacy and distributed learning benefits \cite{wen2023survey}. Therefore, the comparison consider both the performance metrics and the privacy advantages offered by the FL central model.

We also compare our FL model with the KNN approach to see the differences in the evaluation matrix. TABLE~\ref{tab:model_performance} represents a benchmark analysis of FL, CL and KNN approaches. Both FL and CL techniques perform better, however, based on our measurements the KNN is listed with 12.81 meters, while the KNN baseline for UJI is about 8.5 meters, as for example published in Chapter 3 of \cite{torres2017realistic} using the same dataset, however, it is common and normal that results of different implementations might differ. FL will always win if the choice between these approaches depend on factors like data privacy, server reliability, bandwidth efficiency, and scalability. In such scenarios, the KNN algorithm might not be as effective for indoor localization compared to the machine learning techniques used.

\section{Obtained Results and Perspectives} 
\label{sec:results}

\begin{figure}[t]
    \centering
    \includegraphics[scale=0.348]{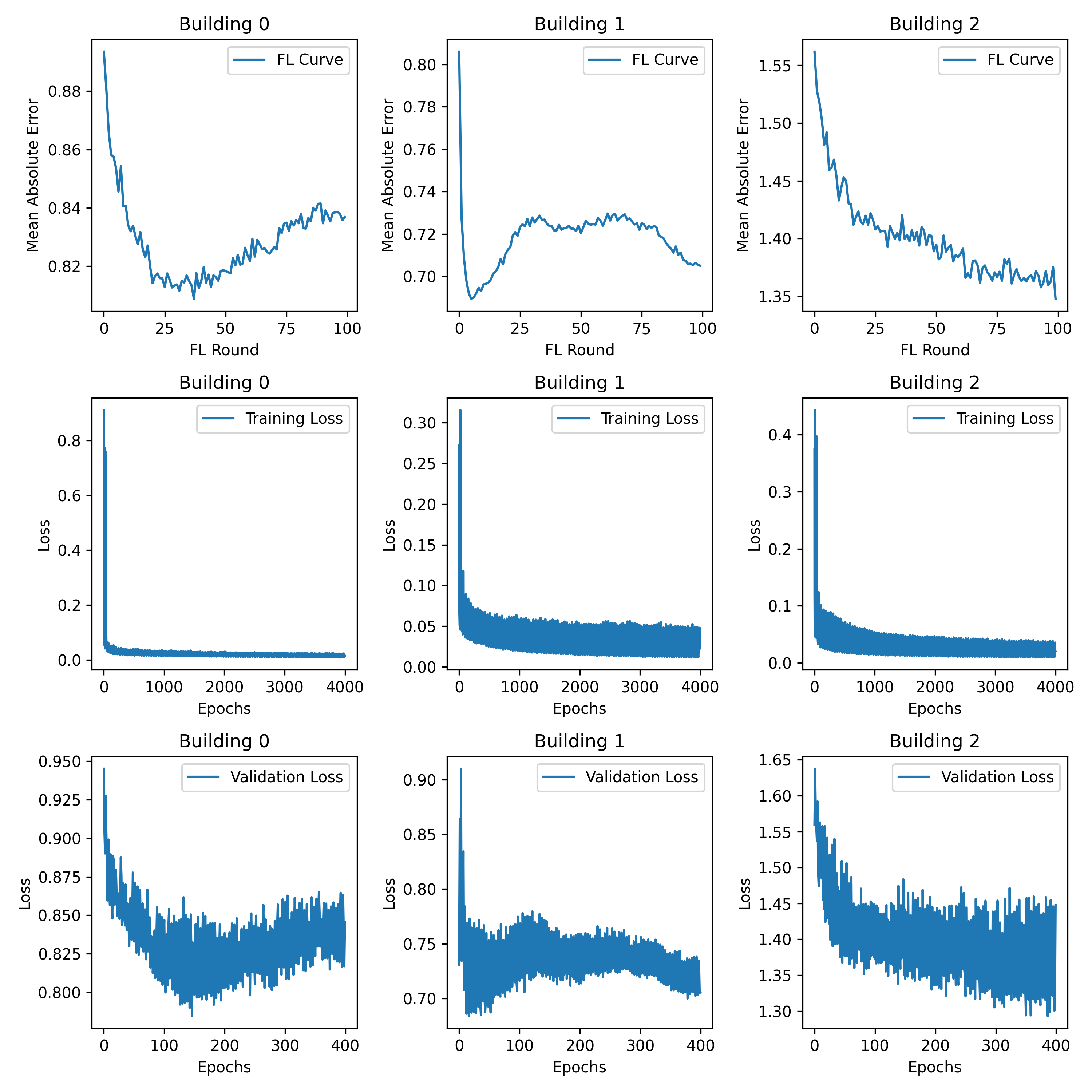}
    \caption{Building Learning Curve and Losses.}
    \label{fig:g}
\end{figure}

In this section we present the results obtained, along with the insights and potential implications these results carry. %This section serves as a critical aspect of our thesis, as it showcases the outcome of our research efforts and lays the foundation for further analysis and discussions.
Fig.~\ref{fig:g} illustrates the outcomes of our code execution for training the Building Models, which comprised 100 rounds and 10 epochs. Each epoch ran for 10 iterations per floor before advancing to the subsequent round. The total number of epochs will be; 

\noindent
\textbf{Training Loss:} 10 epochs per floor * 4 floors per round * 100 rounds = 4000 epochs

\noindent
\textbf{Validation Loss:} 1 evaluation per floor per round * 4 floors per round per building * 100 rounds = 400 epochs

In the early stages of training, the model exhibited a steep decline in both training and validation losses, signifying rapid convergence and learning. As the epochs progressed, the rate of improvement slowed down, but the losses continued to decrease steadily. This indicates the model's continuous adaptation to the data and an enhancement in the overall model's performance up to this stage.

%Notably, Building 0 displayed an improvement in the initial 50 rounds, but this progress was followed by a decline in performance over the next 50 rounds. Conversely, Building 1 and Building 2 demonstrated a consistent enhancement throughout the entire duration. Building 0's initial improvement followed by a subsequent decline suggests a potential over fitting issue or a unique characteristic in the data specific to that building. In contrast, the consistent improvement in Buildings 1 and 2 implies that the model is effectively generalizing patterns and features from the data shared by these buildings.

%To understand the index value of x-axis for the training loss which is 4000 and validation loss which is 400. We are training the local models for 10 epochs per floor in each federated learning round. Since we have 4 floors per building and 10 epochs per floor, that adds up to 40 epochs per building in one FL round. We repeat this process for 100 rounds, which sums up to 4000 epochs per building over all rounds.

%However, we evaluate the validation loss after each local model has been trained on a floor. As there are 4 floors, validation loss is assessed 4 times per round for each building. This process repeats over 100 rounds, leading to a cumulative count of 400 evaluations.

%To understand it further we can break it down here.

%\textbf{Note:} To ensure clarity and improved comprehension, all the plots below will display values in meters. To achieve this, we have converted the previously normalized values back to their original, real-world measurements.

Next we executed the FL central model for 1000 Epochs. % and observed an initial high localization error of approximately 60 meters. However, 
As the training progressed, the model's performance significantly improved, and the localization error decreased to around 10 meters within the first 100 epochs. Upon completion of 1000 epochs we achieve localization error up to 5.45 meters. The validation losses also exhibited a positive trend, with the localization error decreasing to 10.86 meters. Fig. 3 illustrates the localization error for the FL central model. The substantial reduction in localization error during the initial training epochs highlights the effectiveness of the FL central model in improving its accuracy. The decreasing trend of validation losses further supports the model's ability to generalize well on unseen data, contributing to its overall robustness. %Moreover, the consistent decrease in validation losses indicates that the model demonstrates enhanced generalization on unseen data, contributing to its overall robustness and ability to accurately localize objects in real-world coordinates.

To compare this result with the state of the art methods, firstly, we executed the CL central model for 1000 Epochs. % and observed the localization error around 90 meters in the start, in contrast to the FL Central model, which began with a lower error of 60 meters. This suggests that the FL Central Model exhibits better performance right from the early epochs. 
As the training progressed, similar to the FL central model, the CL central model showed a significant improvement during the first 100 epochs. However, what sets it apart is that it continued to demonstrate a steady reduction in localization error over subsequent epochs, showcasing its sustained progress. Fig. 3 provides a comparison of both approaches.

\begin{figure}[h]
    \centering
%{0.39\textwidth}
        \centering

\includegraphics[width=\linewidth]{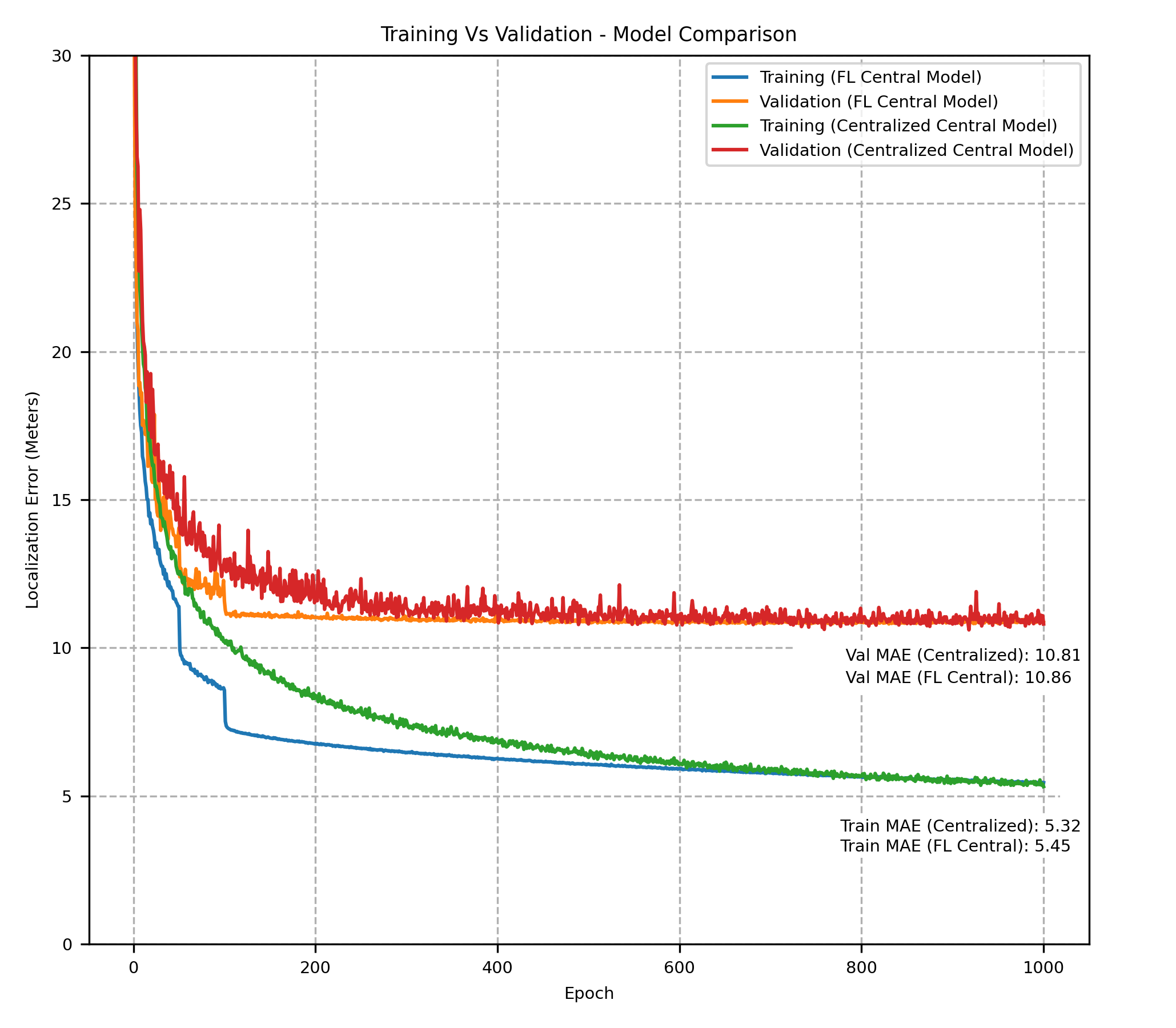}
        \caption{Comparison of FL \& CL Central Models.}
\end{figure}
%\begin{figure}[h]
   % \centering
    %\begin{subfigure}{0.39\textwidth}
       % \centering
        %\includegraphics[width=\linewidth]{common/Images/Training and Validation - FL Central Model.png}
        %\caption{FL Central Model Localization Error.}
        %\label{fig:3a}
    %\end{subfigure}
    %\hfill
%    \begin{subfigure}{0.4\textwidth}
   %     \centering
        %\includegraphics[width=\linewidth]{common/Images/Training and Validation - Centralized Central Model.png}
      %  \caption{CL Central Model Localization Error.}
        %\label{fig:3b}
    %\end{subfigure}
    
    %\begin{subfigure}{0.4\textwidth}
       % \centering
        %\includegraphics[width=\linewidth]{common/Images/Comparison_Models_Localization_Error_Zoomed_with_Last_Values.png}
        %\caption{Comparison of FL \& CL Central Models.}
        %\label{fig:3c}
    %\end{subfigure}
    %\caption{Localization errors of FL and CL Central Models.}
    %\label{fig:3}
%\end{figure}

 %\includegraphics[width=\linewidth, height = 9.5cm]

%Table \ref{Table:c} \& \ref{Table:d} presents a benchmark analysis of our FL central model with the CL central model using the same dataset.

The reason behind the CL central model's superior performance compared to the FL central model lies in its centralized learning approach. In CL, all data is aggregated and processed on a single central server, allowing the model to access a comprehensive dataset, which leads to better model convergence and more robust generalization. On the other hand, the FL approach, employed by the FL central model, involves training models across multiple floors without sharing raw data, potentially resulting in slower convergence and higher variability in performance across devices.

%It is essential to reiterate that both FL and CL Central models were trained using the same training and validation datasets. When examining the training trend of both central models, distinct patterns emerge. 

%In the FL Central model, the validation localization error follows a consistent trajectory from epochs 200 to 1000. The disparity between the training and validation localization error is approximately 5.5 meters, with the training error at 10.86 meters and the validation error hovering at about 5.45 meters.

%Conversely, in the Centralized Central model, both the training and validation trends are not as steady. The training trend slightly improves, displaying a gentle downward trajectory over the epochs, while the validation trend slightly degrades, following a steady response over the same period. This deviation indicates that the model's performance may be more sensitive to variations in the validation data compared to the FL Central model.

%It is evident that both models display consistent trends in their validation localization errors, but with slight differences in the magnitude of the error. The FL Central model exhibits a slightly larger gap between training and validation errors, while the Centralized Central model shows a more closely aligned performance between the two.

\begin{table}[t]
\caption{Benchmark Summary.}
    \centering
    \scalebox{0.9}{
    \begin{tabular}{|c|c|c|c|}
    \hline
    Model & Total Epochs & Training MDE (meters) & Validation MDE (meters) \\
    \hline
    FL & 1000 & 5.45 & 10.86 \\
    \hline
    CL & 1000 & 5.32 & 10.81 \\
    \hline
    KNN & - & - & 12.81 \\
    \hline
    \end{tabular}
    }
    \label{tab:model_performance}
\end{table}

%\begin{table}[ht]
%\centering
%\begin{subtable}{0.45\textwidth}
%\centering
%\scalebox{1}{
%\begin{tabular}{|c|c|c|}
%\hline
%%\rowcolor{yellow}
%\textbf{Model} & \textbf{Total Epochs} & \textbf{MDE (meters)} \\ \hline
%FL Central Model & 1000 & 5.45 \\
%\hline
%CL Central Model & 1000 & 5.32 \\
%\hline
%\end{tabular}
%}
%\caption{Benchmark on UJIndoorLoc training data.}
%\label{Table:c}
%\end{subtable}

%\hfill

%\begin{subtable}{0.45\textwidth}
%\centering
%\scalebox{1}{
%\begin{tabular}{|c|c|c|}
%\hline
%\rowcolor{yellow}
%\textbf{Model} & \textbf{Total Epochs} & \textbf{MDE (meters)} \\ \hline
%FL Central Model & 1000 & 10.86 \\
%\hline
%CL Central Model & 1000 & 10.81 \\
%\hline
%\end{tabular}
%}
%\caption{Benchmark on UJIndoorLoc validation data.}
%\label{Table:d}
%\end{subtable}
%\caption{Benchmark results.}
%\label{Table:ab}
%\end{table}

\section{Conclusion}
\label{sec:Conclusion}

In this paper, we have presented a federated machine learning approach for dynamic indoor environment, prioritizing data privacy. By adapting a DNN architecture we addressed the challenges faced by the traditional centralized machine learning methods such as the privacy concerns, bandwidth limitation and server reliability issues. Our results showed that with only a marginal increase of 3.44\% in localization error compared to the CL model, the FL model perform comparatively similar to the CL model and additionally address the critical concerns of data privacy and bandwidth efficiency showcasing its robustness and generalization capabilities.

Our future research direction includes applying our algorithm to the Visible Light Communication (VLC) data. For this purpose, we have conducted VLC-based measurements at the BOSCH factory in Blaichach, Germany. The results from this experiment will significantly contribute to the advancement of VLC-based indoor localization techniques, which may find applications in various industries, including manufacturing, logistics, and smart buildings.

%Additionally, we envision extending our approach where each floor can be considered a single user, and the same model can be implemented on individual mobile devices. In this scenario, an application on the mobile device processes and trains a model based on self-recorded data, then shares the weights with the local server, which functions as the floor model. This implementation could further enhance the scalability and adaptability of our federated learning approach in diverse environments.


% \begin{acks}
% The acknowledgment goes here!
% \end{acks}

% \bibliographystyle{ACM-Reference-Format}
\bibliographystyle{IEEEtran}
\bibliography{references}

% %%
% %% If your work has an appendix, this is the place to put it.
% \appendix





% Etiam commodo feugiat nisl pulvinar pellentesque.


% Nam id fermentum dui. Suspendisse sagittis tortor a nulla mollis,

\end{document}